\documentclass{article}

\PassOptionsToPackage{numbers, compress}{natbib}

\usepackage[dblblindworkshop,final]{neurips_2026}

\usepackage[utf8]{inputenc} 
\usepackage[T1]{fontenc}    
\usepackage{hyperref}       
\usepackage{url}            
\usepackage{booktabs}       
\usepackage{amsfonts}       
\usepackage{nicefrac}       
\usepackage{microtype}      
\usepackage{xcolor}         

\usepackage{amsmath,amssymb,amsthm}
\usepackage{graphicx}
\usepackage{enumitem}
\IfFileExists{inconsolata.sty}{\usepackage{inconsolata}}{}

\newcommand{\dataset}{FOF-Bench}
\newcommand{\modelname}{FOF-80M}
\newcommand{\basename}{F2LLM}
\newcommand{\disc}{DiSC}
\newcommand{\styleremix}{StyleRemix}
\newcommand{\summeval}{SummEval}
\newcommand{\snli}{SNLI}

\workshoptitle{Linguistic Principles for Foundation Models (LP4FM)}
\title{Focus On Facts: Stylistically Invariant and Factually Sensitive Text Embeddings}

\author{%
  Maximus Adeola \\
  University of Maryland, Baltimore County \\
  \texttt{research@maximusadeola.com} \\
}

\begin{document}
\maketitle

\begin{abstract}
Text embedding models can conflate surface-form similarity with factual identity. Given an altered near-copy with one corrupted fact and a differently worded fact-preserving rewrite, the base model prefers the altered near-copy. We test whether hard-negative contrastive supervision can reduce this surface-similarity shortcut. We construct \dataset{}, an 83,263-triplet benchmark spanning five domains. Each anchor is paired with an LLM-generated, fact-preserving stylistic rewrite and a matched fact-altered near-copy. We fine-tune \modelname{} (\textbf{F}ocus \textbf{O}n \textbf{F}acts), an 80M-parameter embedding model, with a triplet contrastive objective that ranks the rewrite above the factual alteration. On 8,327 held-out test triplets, the base model chooses the factually altered text in 99.58\% of cases (0.42\% factual preference rate; mean separation $-0.1650$). After fine-tuning, mean positive--negative separation is $+0.0329$, 98.5\% of individual margins move in the intended direction, and the preference rate reaches 29.37\%. On \disc{}, an external style-transfer corpus, mean cross-style similarity increases from 0.7905 to 0.9559. The model also penalizes omission and transfers poorly to directional tasks such as NLI and summarization. We characterize \modelname{} as a soft factual fingerprint for studying factual retention under stylistic variation, not as a general-purpose factuality metric.
\end{abstract}

\section{Introduction}

Consider three texts:

\begin{quote}\small
\textbf{Anchor:} We present \textit{ab initio} calculations of the electronic energy loss of charged particles moving outside a magnesium surface. \\

\textbf{Fact-altered near-copy:} We present \textit{ab initio} calculations of the electronic energy loss of charged particles moving outside a beryllium surface. \\

\textbf{Fact-preserving rewrite:} In this highly detailed scientific work, we put forward a set of comprehensive \textit{ab initio} calculations pertaining to the electronic energy loss experienced by charged particles as they travel in close proximity to, but outside of, a magnesium surface.
\end{quote}

The desired behavior is the correct factual ordering: a factual-retention embedding should place the fact-preserving rewrite closer to the anchor than the fact-altered near-copy, even though the altered near-copy shares almost all surface wording. Across thousands of examples, the base model shows the opposite behavior. On our held-out \dataset{} test set, the base model (\texttt{\basename{}-v2-80M}) assigns a mean anchor--positive similarity of 0.8269, but an anchor--negative similarity of 0.9919. This yields a negative mean separation of $-0.1650$ and a factual preference rate of just 0.42\%. When lexical overlap conflicts with factual identity, the base model ranks the factually altered near-copy first in 99.58\% of instances.

\modelname{} reverses this average preference. It raises mean anchor--positive similarity to 0.9823, lowers mean anchor--negative similarity to 0.9495, and changes mean separation from $-0.1650$ to $+0.0329$ (a $+0.1979$ shift; Cohen's $d$ moves from $-2.80$ to $+0.34$). In 98.5\% of held-out triplets, the margin moves in the intended direction. The pairwise factual preference rate increases from 0.42\% to 29.37\%. Hard negatives still outrank positives in most individual comparisons, so \modelname{} does not solve the benchmark. The results show that contrastive training can change the information prioritized by the embedding space.

This paper studies a specific limitation of general-purpose embedding models: factual content, surface form, writing style, syntax, and broad semantic relatedness are all encoded into a single similarity space. This is often desirable for retrieval and clustering, but it makes it difficult to isolate whether two texts preserve the same factual information under changes in wording, length, or style.

We construct synthetic contrastive data that separates factual retention from stylistic presentation. For each source text, an LLM generates a fact-preserving positive rewrite that may be much longer, more rhetorical, or stylistically different. A second prompt generates a negative variant that preserves most wording but introduces one subtle factual or logical alteration. Fine-tuning requires each anchor to rank its stylistically transformed positive above its matched fact-altered near-copy. We use \texttt{codefuse-ai/F2LLM-v2-80M}~\cite{zhang2026f2llmv2inclusiveperformantefficient}, a lightweight embedding model with last-token pooling. We ask two research questions:
\begin{itemize}[nosep]
  \item[\textbf{RQ1}] Can contrastive fine-tuning on fact-preserving rewrites and subtle factual alterations shift an embedding model toward factual-retention behavior?
  \item[\textbf{RQ2}] What capability profile and trade-offs emerge from this training, especially under style transfer, entailment, and summarization?
\end{itemize}

We evaluate \modelname{} across four settings. \dataset{} tests the intended behavior: distinguishing fact-preserving rewrites from near-copy factual alterations. \disc{}, introduced by \styleremix{}~\cite{fisher2024styleremixinterpretableauthorshipobfuscation}, is repurposed as an external style-invariance evaluation. \snli{} and \summeval{} test partial-information behavior: \snli{} combines omission, contradiction, and directional entailment, while \summeval{} tests whether source--summary similarity tracks human judgments under heavy compression.

Our main findings are:
\begin{itemize}[nosep]
  \item \textbf{Fine-tuning shifts the embeddings toward factual retention.} Mean separation changes from $-0.1650$ to $+0.0329$, 98.5\% of triplet margins move in the intended direction, and the positive-ranking rate increases from 0.42\% to 29.37\% (Table~\ref{tab:fofbench-results}).
  \item \textbf{Stylistic invariance improves.} On \disc{}, mean cross-style similarity increases from 0.7905 to 0.9559 (Table~\ref{tab:disc-results}).
  \item \textbf{The model develops a hash-like capability profile.} \modelname{} maps style variants closer together, but also compresses many text pairs into a high-cosine region.
  \item \textbf{Partial-information tasks reveal clear trade-offs.} \modelname{} underperforms the base model on \snli{} and \summeval{}, consistent with an omission penalty.
  \item \textbf{Qualitative analysis reveals entity overpowering.} Dense overlap in proper nouns and topic entities can cause \modelname{} to assign high similarity despite relational errors or hallucinated clauses.
\end{itemize}

\section{Related Work}

\paragraph{Semantic embedding models.}
Early work in supervised sentence representation learning demonstrated that training encoders on NLI yields highly transferrable representations~\cite{conneau2018supervisedlearninguniversalsentence}. Sentence-BERT~\cite{reimers2019sentencebertsentenceembeddingsusing} and its successors (SimCSE~\cite{gao2022simcsesimplecontrastivelearning}, Sentence-T5~\cite{ni2021sentencet5scalablesentenceencoders}, E5~\cite{muennighoff2023mtebmassivetextembedding}) produce fixed-dimensional text representations. Recent general-purpose embedding families include BGE~\cite{xiao2024cpackpackedresourcesgeneral}, GTE~\cite{li2023generaltextembeddingsmultistage}, Nomic Embed~\cite{nussbaum2025nomicembedtrainingreproducible}, Arctic Embed~\cite{merrick2024arcticembedscalableefficientaccurate}, and Qwen embeddings~\cite{zhang2025qwen3embeddingadvancingtext}. These models are optimized for broad semantic usefulness. Recent work has also extended embedding design through adversarial batch clustering and corpus-level contextual conditioning \cite{morris2024contextualdocumentembeddings}, as well as interpretable question-answering embeddings that represent texts via binary LLM judgments \cite{benara2024craftinginterpretableembeddingsasking}. Our goal is different: we study whether a base embedding model can be shifted toward factual-retention behavior under stylistic variation.

\paragraph{Content preservation and factuality metrics.}
Babakov et al.~\cite{babakov-etal-2022-large} evaluated automatic measures for content preservation, finding that NLI-based maximal implication strength correlates best with human judgments. SummEval~\cite{fabbri2021summevalreevaluatingsummarizationevaluation} and Kry\'sci\'nski et al.~\cite{kryscinski2019evaluating} focus on evaluation metrics rather than representation learning. Our approach is complementary: we investigate whether factual-retention behavior can be shaped directly inside an embedding space.

\paragraph{Style invariance and authorship obfuscation.}
StyleRemix~\cite{fisher2024styleremixinterpretableauthorshipobfuscation} uses LoRA modules to perturb specific style axes and measures content preservation via Sentence Transformers. We use their \disc{} dataset as an external style-invariance evaluation. Zhao et al.~\cite{10.1145/3774904.3792669} propose counterfactual reasoning for cross-style hate speech detection. These approaches modify the generation space; we modify the embedding space directly.

\paragraph{Positioning.}
Semantic similarity models encode style and content in the same representation, while content-preservation metrics evaluate factuality without changing the representation. We use contrastive training to shift an embedding model toward factual retention.

\section{Method}\label{sec:method}

\subsection{Dataset Construction}\label{sec:method-data}

\paragraph{Source texts.}
We construct \dataset{}, a synthetic factual-retention dataset, normalize the text, and fine-tune all parameters of a lightweight last-token-pooled embedding model with a contrastive objective. We begin with 83,689 snippets drawn from five sources (Table~\ref{tab:source-counts}). Each source is capped at 20,000 snippets except ReClor, which contributes 3,689 contexts~\cite{yu2020reclorreadingcomprehensiondataset}. Snippets are filtered to approximately 50--500 words, with a mean length of roughly 115 words. We sample from Wikipedia~\cite{wikidump}, arXiv abstracts~\cite{clement2019arxiv}, PubMed bioinformatics abstracts~\cite{gupta_2026_pubmed_bioinformatics_all_years,Sayers2022-wl}, ROCStories~\cite{mostafazadeh-etal-2016-corpus}, and ReClor.

\begin{table}[t]
\centering
\caption{Source composition of the initial snippet pool.}
\label{tab:source-counts}
\begin{tabular}{lr}
\toprule
Source & Snippets \\
\midrule
Wikipedia & 20,000 \\
arXiv & 20,000 \\
PubMed & 20,000 \\
ROCStories & 20,000 \\
ReClor & 3,689 \\
\midrule
Total & 83,689 \\
\bottomrule
\end{tabular}
\end{table}

\paragraph{Fact-preserving positives.}
For each source text, we prompt \texttt{mistral-small-2603} to generate a fact-preserving rewrite. The rewrite may alter style, verbosity, sentence structure, or rhetorical framing, but is instructed not to add, remove, reorder, or contradict factual information. The expansion factor is drawn uniformly from $\{1.5, 2.0, \dots, 5.0\}$.

\paragraph{Fact-altered negatives.}
For the same source text, we prompt the LLM to generate a near-copy containing one subtle factual or logical alteration. Alteration techniques include entity substitution, numeric modification, direction changes, quantifier changes, cause--effect swaps, conditional reversals, and event-order changes.

\paragraph{Generation and cleaning.}
The generation and cleaning pipeline yielded a final cleaned dataset of 83,263 triplets, split into 66,610 training, 8,326 validation, and 8,327 held-out test triplets. Table~\ref{tab:wordstats} reports word count statistics. Table~\ref{tab:example-triplet} shows an example.

\paragraph{Context length and truncation.} Positives average 701 tokens, and 16.8\% exceed the model's 1,024-token context limit; fewer than 0.03\% of anchors and negatives exceed it. Longer inputs are truncated during encoding. Section~\ref{sec:benchmark-fofbench} reports results separately for truncated and untruncated positives.

\begin{table}[t]
\centering
\caption{Word count statistics for the three fields in the cleaned dataset (83,263 triplets).}
\label{tab:wordstats}
\begin{tabular}{lrrrrr}
\toprule
Field & Mean & P95 & P99 & Max \\
\midrule
Anchor (original) & 116.2 & 252 & 311 & 500 \\
Positive (expanded) & 573.8 & 1,160 & 1,644 & 6,020 \\
Negative (altered) & 116.3 & 252 & 311 & 500 \\
\bottomrule
\end{tabular}
\end{table}

\begin{table}[t]
\centering
\caption{Example triplet from the training data (ROCStories source).}
\label{tab:example-triplet}
\begin{tabular}{lp{11.7cm}}
\toprule
\textbf{Anchor} & My sister-in-law loves Bubble Witch. In order to get a better score she needed to get FB friends. She issued friend requests to a lot of strangers playing this game. She ended up with a lot of friends from France and Germany. On her birthday she received many greetings in French and German. \\
\textbf{Positive} & My sister-in-law is absolutely fond of playing the popular mobile game, Bubble Witch. In her quest to achieve a higher score in the game, she realized that she would need to expand her circle of friends on Facebook. Consequently, she went ahead and sent out friend requests to quite a few individuals she did not know personally, all of whom were also engaged in playing this particular game. As a result of this strategy, she managed to accumulate numerous friends who reside in countries like France and Germany. When her birthday finally arrived, she was pleasantly surprised to receive a heartfelt number of greetings and well-wishes, most of which were written in French and German. \\
\textbf{Negative} & My sister-in-law loves \emph{Candy Crush}. In order to get a better score she needed to get FB friends. She issued friend requests to a lot of strangers playing this game. She ended up with a lot of friends from France and Germany. On her birthday she received many greetings in French and German. \\
\bottomrule
\end{tabular}
\end{table}

\subsection{Text Normalization}\label{sec:normalization}

All final training and evaluation runs use normalized text. We apply Unicode normalization, replace typographic quotes with ASCII equivalents, remove non-ASCII artifacts, strip control characters, and collapse repeated whitespace. In early experiments, arXiv-derived anchors contained PDF-parsing line breaks that the LLM often removed from positives and negatives. Normalization prevents the model from using these formatting differences as shortcuts.

\subsection{Model Architecture and Training}\label{sec:model-training}

We fine-tune all parameters of \texttt{codefuse-ai/F2LLM-v2-80M}~\cite{zhang2026f2llmv2inclusiveperformantefficient}, an 80M-parameter Qwen3-based model~\cite{zhang2025qwen3embeddingadvancingtext} with last-token pooling, 320-dimensional embeddings, and a 1,024-token context limit. We chose last-token pooling because it uses one contextualized state for the sequence, whereas mean pooling may dilute a small factual change across token states.

We train with \texttt{CachedMultipleNegativesRankingLoss}. Each example is a triplet $(A_i, P_i, N_i)$ containing an anchor, a fact-preserving positive, and its matched factual negative. Other positives and negatives in the batch provide additional candidates. Let $\mathcal{C}_i = \{P_i,N_i\} \cup \{P_j,N_j\}_{j \ne i}$ and let $s$ denote cosine similarity. The loss is
{\small
\begin{equation}
\mathcal{L}_i = -\log
\frac{\exp(s(A_i,P_i)/\tau)}
{\sum_{D \in \mathcal{C}_i}\exp(s(A_i,D)/\tau)} .
\end{equation}
}
We use $\tau=0.05$ (scale factor 20.0) and train for 3 epochs with per-device batch size 16, learning rate $5\times10^{-5}$, cosine decay, 10\% warmup, weight decay 0.01, and cached-loss mini-batch size 32. Training and experimentation were conducted on an NVIDIA RTX Pro 6000 GPU via Google Colab and took approximately 20--30 hours.

\paragraph{Model selection.} We saved checkpoints every 100 steps and selected one using only the validation split. Successive halving evaluated all checkpoints on 300 validation triplets, the top 15 on 1,500, and the top 3 on all 8,326. Checkpoint-12492 had the highest validation preference rate and was then evaluated on the held-out test split.

\subsection{Evaluation Metrics}\label{sec:metrics}

For \dataset{}, let $m_i = \cos(A_i, P_i) - \cos(A_i, N_i)$. We report the pairwise factual preference rate ($\Pr(m_i > 0)$), mean margin ($m = \frac{1}{N}\sum m_i$), the per-example margin shift ($\Delta m_i = m_i^{\text{FOF}} - m_i^{\text{Base}}$), Cohen's $d$, and ROC-AUC. For \disc{}, we compute mean cross-style similarity, factual-cluster silhouette score, style-distractor accuracy, and style-agnostic retrieval R-precision. For \snli{}, we train logistic regression and XGBoost classifiers on embedding-pair features $[\mathbf{u}, \mathbf{v}, |\mathbf{u}-\mathbf{v}|]$. For \summeval{}, we compute Spearman correlation between source--summary cosine similarity and expert human annotations.

\section{Experiments}

\subsection{Evaluation Suite}\label{sec:eval-suite}

Our evaluation characterizes the effects of the fine-tune. \dataset{} tests whether the model distinguishes fact-preserving rewrites from near-copy factual alterations. \disc{}, introduced by \styleremix{}, provides an external style-invariance evaluation. \snli{} and \summeval{} test behavior when one text contains only part of the information in another. We compare \modelname{} with its base model, \texttt{codefuse-ai/F2LLM-v2-80M}, to isolate the effect of fine-tuning.

\subsection{\dataset{}: Factual Retention}\label{sec:benchmark-fofbench}

\subsubsection{Setup.}
The \dataset{} test set comprises 8,327 triplets. We report factual preference rate, mean anchor--positive and anchor--negative cosine similarity, mean separation, Cohen's $d$, and ROC-AUC.

\subsubsection{Results.}

\begin{table}[t]
\centering
\small
\setlength{\tabcolsep}{3pt}
\caption{Main \dataset{} results on the held-out test set (8,327 triplets).}
\label{tab:fofbench-results}
\begin{tabular}{lrrr}
\toprule
Metric & Base (\basename{}) & \modelname{} & $\Delta$ \\
\midrule
Pairwise Preference Rate (\%) & 0.42 & \textbf{29.37} & $+28.95$ \\
Anchor--Positive $\cos$ & 0.8269 & \textbf{0.9823} & $+0.1554$ \\
Anchor--Negative $\cos$ & 0.9919 & \textbf{0.9495} & $-0.0424$ \\
Mean Separation & -0.1650 & \textbf{+0.0329} & $+0.1979$ \\
Cohen's $d$ & -2.7966 & \textbf{+0.3397} & $+3.1363$ \\
ROC-AUC & 0.0070 & \textbf{0.4002} & $+0.3932$ \\
\bottomrule
\end{tabular}
\end{table}

The base model assigns the fact-altered near-copy higher similarity in 99.58\% of triplets, producing a preference rate of 0.42\% and ROC-AUC 0.0070 (Table~\ref{tab:fofbench-results}). The near-copy's lexical overlap therefore outweighs factual identity in almost every triplet. After fine-tuning, anchor--positive similarity rises from 0.8269 to 0.9823 and anchor--negative similarity falls from 0.9919 to 0.9495. Mean separation changes from $-0.1650$ to $+0.0329$, a shift of $+0.1979$, and Cohen's $d$ changes from $-2.80$ to $+0.34$. In 98.5\% of held-out triplets, the margin moves in the intended direction ($\Delta m_i > 0$). The pairwise preference rate increases from 0.42\% to 29.37\%. Since the rate remains below 50\%, hard negatives still rank above positives in most triplets. Because fine-tuning changes cosine calibration, we compare rankings and margins within each model rather than absolute cosine values across models.

\paragraph{Length-confound and truncation analysis.}
Because positives are stylistically expanded relative to anchors and negatives, we test whether FOF-80M's gains are driven by positive verbosity. In the base model, triplet margins strongly anti-correlate with positive expansion ratio ($r = -0.425, \rho = -0.535$), dropping from $-0.1194$ in the lowest expansion quartile ($<3.56\times$) to $-0.2165$ in the highest ($\ge 6.70\times$). After fine-tuning, FOF-80M eliminates this expansion penalty without introducing a positive length bias: across all 8,327 test triplets, margin correlation with expansion ratio is near zero ($r = -0.021, \rho = -0.057$) and slightly negative with raw word count ($r = -0.080, \rho = -0.172$). On the 6,932 untruncated test triplets whose positives fit within the 1,024-token limit ($L(P) \le 1024$), FOF-80M obtains a 30.2\% preference rate and $+0.0377$ mean margin, with near-zero correlation against both word count ($r = +0.009$) and expansion ratio ($r = +0.006, \rho = -0.020$). Moreover, shorter positives in the 100--250 word ($N=870$) and 250--500 word ($N=3{,}173$) bins achieve the strongest margins ($+0.0422$ and $+0.0397$; preference rates 30.5\% and 31.8\%), whereas the 1,395 truncated positives ($L(P) > 1024$) drop to 22.9\%. Thus, while binary style differences between unexpanded negatives and rewritten positives remain a structural property of the benchmark (Section~6.3), within-distribution length variation shows no evidence that longer outputs receive an artificial advantage.

\subsection{\disc{}: Stylistic Invariance}\label{sec:benchmark-disc}

\subsubsection{Setup.}
\disc{} was introduced by \styleremix{} as a dataset of texts rewritten along multiple style directions~\cite{fisher2024styleremixinterpretableauthorshipobfuscation}. We repurpose it as an external style-invariance evaluation. We measure:
\begin{enumerate}[nosep]
  \item \textbf{Mean stylistic invariance}: mean cosine similarity across same-content style pairs.
  \item \textbf{Factual silhouette score}: clustering quality using source identity as ground truth~\cite{ROUSSEEUW198753}.
  \item \textbf{Style-distractor accuracy}: does the model rank same-content/different-style above different-content/different-style?
  \item \textbf{Style-agnostic retrieval R-precision}: for a corpus of texts across many facts and styles, retrieve the top-$k$ neighbors and measure how many match the correct source.
\end{enumerate}

\subsubsection{Results.}

\begin{table}[t]
\centering
\small
\setlength{\tabcolsep}{3pt}
\caption{Stylistic invariance results on \disc{}. \disc{} is repurposed from \styleremix{} as an external style-invariance evaluation.}
\label{tab:disc-results}
\begin{tabular}{lrrr}
\toprule
Metric & Base & \modelname{} & $\Delta$ \\
\midrule
Mean Stylistic Invariance & 0.7905 & \textbf{0.9559} & $+0.1654$ \\
Factual Silhouette Score & 0.6843 & \textbf{0.7147} & $+0.0304$ \\
Style-Distractor Acc. (\%) & \textbf{99.87} & 98.93 & $-0.94$ \\
Style-Agnostic R-Prec. (\%) & \textbf{98.84} & 88.36 & $-10.48$ \\
\bottomrule
\end{tabular}
\end{table}

\modelname{} increases mean cross-style similarity from 0.7905 to 0.9559 and factual silhouette score from 0.6843 to 0.7147 (Table~\ref{tab:disc-results}). Style-agnostic R-precision drops from 98.84\% to 88.36\%. The model maps style variants closer together but loses some nearest-neighbor discrimination in large retrieval pools.

\subsection{\snli{}: Logical Generalization}\label{sec:benchmark-snli}

\subsubsection{Setup.}
We train logistic regression and XGBoost classifiers on embedding features $[\mathbf{u}, \mathbf{v}, |\mathbf{u}-\mathbf{v}|]$ from the \snli{} validation set (9,842 pairs) and evaluate on the test set (9,824 pairs)~\cite{bowman2015largeannotatedcorpuslearning}. We evaluate both 3-way (entailment, neutral, contradiction) and 2-way (entailment vs.\ contradiction, dropping neutral) classification, and report raw cosine similarity per class.

\subsubsection{Results.}

\begin{table}[t]
\centering
\caption{Full \snli{} evaluation for Base and \modelname{} (9,824 test pairs).}
\label{tab:snli-results}
\begin{tabular}{lrr}
\toprule
Metric & Base & \modelname{} \\
\midrule
\multicolumn{3}{l}{\textbf{Raw cosine similarity}} \\
\quad Entailment mean $\cos$ & 0.5231 & 0.8890 \\
\quad Neutral mean $\cos$ & 0.4344 & 0.8582 \\
\quad Contradiction mean $\cos$ & 0.2992 & 0.8042 \\
\quad Separation (Ent $-$ Con) & \textbf{0.2239} & 0.0848 \\
\midrule
\multicolumn{3}{l}{\textbf{3-way classification}} \\
\quad Logistic Regression (\%) & \textbf{71.07} & 52.91 \\
\quad XGBoost (\%) & \textbf{67.02} & 52.15 \\
\midrule
\multicolumn{3}{l}{\textbf{2-way binary (entailment vs con)}} \\
\quad Logistic Regression (\%) & \textbf{87.49} & 70.43 \\
\quad XGBoost (\%) & \textbf{83.79} & 70.60 \\
\bottomrule
\end{tabular}
\end{table}

\snli{} exposes a mismatch between factual retention and natural language inference (Table~\ref{tab:snli-results}). Let $F(T)$ represent the factual propositions expressed by text $T$. Entailment is directional: hypothesis $T_2$ is entailed by premise $T_1$ when $F(T_2) \subseteq F(T_1)$. For example, ``A person in an orange shirt reaching up'' entails ``A person is reaching up,'' although the hypothesis omits the shirt and its color. \modelname{} is trained for symmetric factual parity, $F(T_1) = F(T_2)$, so it penalizes missing, added, and altered information. Under \modelname{}, all three classes move into a high-similarity region, and entailment--contradiction separation falls from 0.2239 to 0.0848. Probe accuracy falls from 71\% to 53\% for the 3-way task and from 87\% to 70\% for the 2-way task. These results show that the learned representation is poorly suited to directional entailment without further training or calibration.

\subsection{\summeval{}: Summarization Under Heavy Omission}\label{sec:benchmark-summeval}

\subsubsection{Setup.}
We use \summeval{} to test source--summary similarity when omission is expected. The dataset contains 1,600 summaries from 100 CNN/DM articles and 16 systems, with expert annotations~\cite{fabbri2021summevalreevaluatingsummarizationevaluation}. We compute Spearman correlation between source--summary cosine similarity and the expert ratings.

\subsubsection{Results.}

\begin{table}[t]
\centering
\caption{Spearman $\rho$ of source--summary cosine similarity with expert annotations on \summeval{} (1,600 entries).}
\label{tab:summeval-results}
\begin{tabular}{lrr}
\toprule
Human Metric & Base & \modelname{} \\
\midrule
Consistency & \textbf{0.2061} & 0.0198 \\
Relevance & \textbf{0.4124} & 0.1204 \\
Coherence & \textbf{0.2651} & 0.0459 \\
Fluency & \textbf{0.1186} & 0.0128 \\
\bottomrule
\end{tabular}
\end{table}

On \summeval{}, the correlation between source--summary cosine and expert consistency falls from 0.2061 for the base model to 0.0198 for \modelname{} (Table~\ref{tab:summeval-results}). Summaries omit much of the source, while \modelname{} is trained to reward near-complete factual retention. It may therefore penalize a faithful summary for preserving only a subset of the source facts. Together with the \snli{} results, this drop indicates that the model does not handle omission well.

\section{Qualitative Behavioral Analysis}\label{sec:qualitative}

Quantitative metrics show the direction of change, but not the mechanism. We therefore inspect high-disagreement \summeval{} examples where the base model and \modelname{} assign very different source--summary similarities. We organize these examples around four behaviors: omission penalty, entity overpowering, relational hallucination detection, and recognition of fact-dense outliers.

\subsection{Omission Penalty}\label{sec:omit}

In several cases, \modelname{} assigns low similarity to summaries that human annotators judge as fully consistent. For example, a Chelsea--Yoshinori Muto summary receives expert consistency 5.0/5.0 and high base-model similarity (0.9102), but \modelname{} assigns only 0.4492. The summary accurately preserves the main facts, but the article contains additional details that the summary omits. \modelname{} penalizes this missing information despite the summary being faithful. We call this behavior the \emph{omission penalty}: because positives in the training data preserve all the source facts, the model learns to treat missing information as evidence against equivalence. This is useful for strict factual-retention comparison, but harmful for summarization and entailment, where partial omission is expected.

\subsection{Entity Overpowering}\label{sec:entity}

A second failure mode is entity overpowering. In one low-consistency \summeval{} example, a summary combines entities from a Liverpool article---Steven Gerrard, Brendan Rodgers, Raheem Sterling, Anfield---with hallucinated or corrupted claims involving Sterling and drug use. Human experts assign low consistency (2.0/5.0). The base model assigns low similarity (0.4102), but \modelname{} assigns high similarity (0.8945). This suggests that entity overlap can outweigh incorrect relations when both texts remain on the same topic.

\subsection{Relational Hallucination Detection}\label{sec:relational}

In other cases, \modelname{} assigns lower similarity than the base model to relational hallucinations. In a Filipe Luis example, the summary incorrectly states: ``I don't think I'm a player for Atletico Madrid.'' The source article instead reports that Luis wants to stay at Chelsea while Atletico Madrid considers bringing him back. Human experts assign low consistency (2.0/5.0). The base model assigns similarity 0.8711, while \modelname{} assigns 0.6367. This indicates that \modelname{} can penalize certain relation-level mismatches, especially when the hallucinated relation changes the factual role or stance of a named entity.

\subsection{Fact-Dense Outlier Recognition}\label{sec:outlier}

\modelname{} also recognizes some faithful summaries that the base model scores poorly. In a Leicester City example, the summary states that the Foxes host Swansea, sit three points from safety, recently beat West Brom 3--2, and received an injury-time winner from Jamie Vardy. Human experts assign consistency 5.0/5.0. The base model assigns low similarity (0.5352), while \modelname{} assigns high similarity (0.9375). This suggests that \modelname{} can identify compact factual overlap even when the summary omits background framing or differs in style.

\section{Discussion}\label{sec:discussion}

\subsection{Synthesis: Answering the Research Questions}\label{sec:synthesis}

\textbf{RQ1: Can contrastive fine-tuning shift an embedding model toward factual-retention behavior?}
Yes. On \dataset{}, the base model chooses the factually altered near-copy in 99.58\% of instances. After fine-tuning, mean separation changes from $-0.1650$ to $+0.0329$, 98.5\% of test margins move in the intended direction, and the pairwise factual preference rate increases from 0.42\% to 29.37\%. The desired ordering still occurs in fewer than half of individual triplets, but the results show that fine-tuning can shift the embeddings toward factual retention.

\textbf{RQ2: What capability profile and trade-offs emerge?}
\modelname{} improves stylistic invariance on \disc{} but compresses many text pairs into a high-cosine region. It performs worse than the base model on the partial-information tasks \snli{} and \summeval{}, consistent with the omission penalty. Qualitative analysis also finds entity overpowering: overlap in named entities can outweigh relational errors. \modelname{} is therefore an exploratory factual-retention embedding, not a calibrated factuality metric.

\subsection{A Soft Hashing Interpretation}\label{sec:hashing}

We interpret \modelname{} as a soft factual fingerprint rather than a conventional semantic-similarity embedding. It maps stylistic variants into a shared high-similarity region while making some factual substitutions affect their relative ranking. This behavior helps with rhetorical and stylistic variation (Section~\ref{sec:benchmark-disc}) but has three observed costs: omitted facts are penalized (Section~\ref{sec:omit}), entity overlap can outweigh relational errors (Section~\ref{sec:entity}), and unrelated or partially related texts may occupy the same high-cosine region (Section~\ref{sec:benchmark-snli}). \modelname{} is therefore a rank-sensitive factual-retention signal, not a calibrated factuality score.

\subsection{Limitations}\label{sec:limitations}

\paragraph{Incomplete reordering.} Although the average margin becomes positive and 98.5\% of margins move in the intended direction, the pairwise preference rate is 29.37\%. Hard lexical negatives still rank first in most triplets. The benchmark is not solved.

\paragraph{Cosine compression.} The fine-tuned model maps many text pairs into a high-cosine region. This means raw cosine values are poorly calibrated across models; relative ordering and separation are more informative than absolute similarity.

\paragraph{Omission penalty and entity overpowering.} The model penalizes missing information, which harms tasks where partial omission is expected, including natural language inference and summarization. Dense overlap in named entities or topic-specific terms can cause the model to assign high similarity despite relational errors or hallucinated clauses.

\paragraph{Synthetic data quality.} The dataset is LLM-generated and may contain malformed positives, weak negatives, hallucinated expansions, or accidental factual changes. We do not quantify these generation errors with human validation.

\paragraph{Length and stylistic confounding.} Because positive rewrites are stylistically expanded (mean 573.8 words) while negative near-copies preserve the anchor length (mean 116.3 words), the contrastive objective could partially rely on length or generation-style cues alongside factual parity. Future iterations of FOF-Bench should incorporate length-matched stylistic negatives and larger human-audited evaluation subsets as stricter controls.

\paragraph{Single architecture and baseline scope.} We compare only with the direct base model. This isolates the effect of fine-tuning but does not establish performance relative to other embedding models. The experiments use one lightweight last-token-pooled base model.

\subsection{Future Work}\label{sec:future}

\paragraph{Curriculum learning and graded negatives.} The current dataset mostly uses hard binary contrasts: fact-preserving rewrite versus subtle factual alteration. Future work should construct graded negatives with multiple levels of factual change: omission of minor details, loss of specific entities, alteration of relations, and corruption of logical order. This may smooth the current hash-like behavior and teach the model to distinguish degrees of factual deviation.

\paragraph{Higher-quality data, architectures, and training methods.} Human validation or automated filtering with strong factuality models could remove malformed positives, weak negatives, and LLM generation artifacts. Augmenting the dataset with diverse prompts and LLM generators, rather than relying on a single model, may also reduce prompt and model dependent artifacts. Future experiments can test the objective across model sizes, pooling methods, and single- versus multi-stage training.

\section{Conclusion}

We introduced \modelname{}, an embedding model trained with matched factual hard negatives. On \dataset{}, the base model chooses fact-altered near-copies over stylistic, fact-preserving rewrites in 99.58\% of instances. After fine-tuning, mean separation changes from $-0.1650$ to $+0.0329$, 98.5\% of test margins move in the intended direction, and the factual preference rate increases from 0.42\% to 29.37\%.

The model is not a general-purpose factuality metric. It performs worse than the base model on \snli{} and \summeval{}, and qualitative analysis identifies omission penalty and entity overpowering as failure modes. We characterize \modelname{} as a soft factual fingerprint for studying factual retention under stylistic variation, not as a calibrated factuality metric. We release the FOF-Bench dataset at \url{https://huggingface.co/datasets/MaxDevv/FOF-Bench} and the FOF-80M model weights at \url{https://huggingface.co/MaxDevv/FOF-80M}.

\bibliographystyle{plainnat}
\bibliography{refs}

\end{document}